\documentclass[10pt,twocolumn,letterpaper]{article}
\usepackage{cvpr}
\usepackage{amsthm}

\usepackage{enumitem}
\usepackage{booktabs}
\usepackage[table]{xcolor}
\usepackage{pifont}
\newcommand{\cmark}{\ding{51}}

\usepackage{booktabs}
\usepackage{amsfonts}
\usepackage{amsthm}
\usepackage{amsmath}

\usepackage{amssymb}
\usepackage{nicefrac}
\usepackage{microtype}
\usepackage{xcolor}
\usepackage{xspace}
\usepackage{mathtools}
\usepackage{subcaption}
\usepackage[safe]{tipa}
\usepackage{color, colortbl}
\usepackage{bm}
\usepackage{algorithm}
\usepackage{algpseudocode}
\usepackage{multirow}
\usepackage{makecell}
\usepackage{newfloat}
\usepackage{listings}
\DeclareCaptionStyle{ruled}{labelfont=normalfont,labelsep=colon,strut=off}
\floatstyle{ruled}
\newfloat{listing}{tb}{lst}{}
\floatname{listing}{Listing}
\usepackage{listings}

\definecolor{cvprblue}{rgb}{0.21,0.49,0.74}
\usepackage[pagebackref,breaklinks,colorlinks,allcolors=cvprblue]{hyperref}
\def\paperID{}
\def\confName{CVPR}
\def\confYear{2026}
\title{Cluster-Aware Neural Collapse Prompt Tuning for Long-Tailed Generalization of Vision-Language Models}
\author{
Boyang Guo$^{1,2}$\thanks{This work is done during the intern in VIPL group, ICT, CAS.}\ \quad
Liang Li$^{2}$\thanks{Corresponding author: Liang Li (liang.li@ict.ac.cn).} \quad
Lin Peng$^{3}$ \quad
Yuhan Gao$^{1}$ \quad
Xichun Sheng$^{4}$ \quad
Chenggang Yan$^{1, 5}$\\[0.2em]
$^{1}$ Hangzhou Dianzi University, Hangzhou, China\\
$^{2}$ Institute of Computing Technology, Chinese Academy of Sciences, Beijing, China\\
$^{3}$ The Hong Kong Polytechnic University, Hong Kong, China\\
$^{4}$ Macao Polytechnic University, Macao, China\\
$^{5}$Zhejiang Provincial Key Laboratory of Low Altitude Ubiquitous Networking Technology, HDU \\[0.3em]
{\tt\small \{boyang.guo, yuhangao, cgyan\}@hdu.edu.cn, liang.li@ict.ac.cn}\\
{\tt\small p2314922@mpu.edu.mo, noreen.peng@connect.polyu.hk}
}
\begin{document}
\maketitle
\begin{abstract}
Prompt learning has emerged as an efficient alternative to fine-tuning pre-trained vision-language models (VLMs).
Despite its promise, current methods still struggle to maintain tail-class discriminability when adapting to class-imbalanced datasets.
In this work, we propose cluster-aware neural collapse prompt tuning (CPT), which enhances the discriminability of tail classes in prompt-tuned VLMs without sacrificing their overall generalization.
First, we design a cluster-invariant space by mining semantic assignments from the pre-trained VLM and mapping them to prompt-tuned features.
This computes cluster-level boundaries and restricts the constraints to local neighborhoods, which reduces interference with the global semantic structure of the pre-trained VLM.
Second, we introduce neural-collapse–driven discriminability optimization with three losses: textual Equiangular Tight Frame (ETF) separation loss, class-wise convergence loss, and rotation stabilization loss.
These losses work together to shape intra-cluster geometry for better inter-class separation and intra-class alignment.
Extensive experiments on 11 diverse datasets demonstrate that CPT outperforms SOTA methods, with stronger performance on long-tail classes and good generalization to unseen classes.
\end{abstract}
\section{Introduction}
Vision-language models such as CLIP \cite{Radford2021CLIP} and EVA-CLIP \cite{sun2023evaclip} align images and text in a shared embedding space through large-scale contrastive pre-training, and they show strong zero-shot recognition across many downstream datasets.
However, adapting these models to specific tasks is still challenging.
Full fine-tuning is expensive and may degrade the generalization ability of the pre-trained backbone.
Prompt tuning \cite{Yao_2023_CVPR, Yao_2024_CVPR, Li_2025_ICLR, Khattak_2023_ICCV,Yao_2025_TPAMI,Guo_2025_CVPR} has emerged as a compelling alternative.
Instead of updating the entire model, it learns a small number of prompt tokens for the vision and text encoders, which efficiently preserves transferability.
Despite this progress, prompt tuning fails to sufficiently represent tail classes on class-imbalanced data \cite{shi2024efficientlongtailedgeneralizationpretrained}.
Due to its reliance on a small number of learnable parameters, prompt tuning is especially sensitive to class imbalance, where head classes dominate the optimization and tail classes with few examples remain difficult to distinguish.
Bridging the tail-class discriminability gap on real-world long-tailed data remains a major challenge for adapting pre-trained VLMs.
\begin{figure*}[t]
    \centering
    \includegraphics[width=\textwidth]{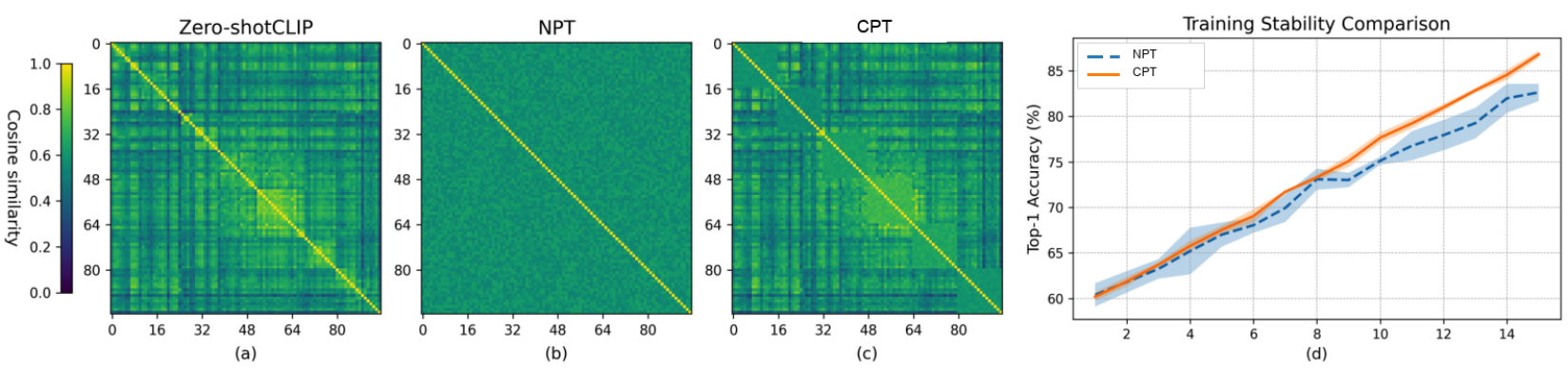}
    \caption{
(a-c): Cosine-similarity matrices of textual features of 96 categories from ZeroshotCLIP, NPT, and CPT.
CPT applies the ETF constraint only within clusters while keeping variation between clusters.
These results show that improving geometric separability can reduce semantic flexibility, while CPT offers a better trade-off by enhancing local separability without collapsing global semantics.
(d): Training stability on EuroSAT (15 epochs, 5 seeds). The shaded area represents ±1 standard deviation. CPT shows significantly lower seed-to-seed variance than NPT.
}
    \label{fig:into}
\end{figure*}
To improve tail-class discriminability, recent methods \cite{didi2024NCprompt, sedov2024exploringembeddingpriorsprompttuning, cho2023distribution} draw inspiration from Neural Collapse (NC).
NC \cite{Papyan_2020, kothapalli2023neuralcollapsereviewmodelling} is a phenomenon observed in the terminal phase of supervised training: features within each class concentrate around a single prototype, and these prototypes form an Equiangular Tight Frame (ETF), i.e., they become maximally separated with uniform pairwise angles.
Some prompt tuning methods enforce a global ETF structure on the textual prototypes of all classes in order to enlarge the geometric margins between classes.
This increases inter-class angular separation and improves discrimination of rare classes.
However, we find that relying on a single global ETF has two main limitations.
First, it distorts the high-level semantic geometry learned by the VLM.
Pre-trained VLMs encode rich hierarchical structure, where semantically related classes remain closer than unrelated ones.
A strict global ETF flattens this hierarchy by forcing all off-diagonal similarities to become nearly identical.
This compresses the effective rank of the similarity matrix, which harms cross-dataset and out-of-distribution (OOD) generalization \cite{harun2025controllingneuralcollapseenhances, hui2022limitationsneuralcollapseunderstanding}.
We visualize this effect in Fig.~\ref{fig:into}(a–c): global ETF drives the prototypes toward a single simplex, while semantic clusters that exist in the pre-trained model disappear.
Second, global ETF constraints enforce fixed angular separations but lack control over absolute orientations, so the constraints are incomplete and training becomes unstable.
Fig.~\ref{fig:into}(d) illustrates that the performance of the SOTA method varies greatly across random seeds.
Overall, existing prompt tuning methods still struggle to achieve both strong tail-class discriminability and robust cross-dataset generalization.
Motivated by these limitations, we propose cluster-aware neural collapse prompt tuning (CPT), a framework that enhances the discriminability of tail classes in prompt-tuned VLMs while preserving their overall generalization.
Specifically, we first design a cluster-invariant space by mining semantic assignments from the pre-trained VLM and mapping them to prompt-tuned features.
This computes cluster-level boundaries and restricts the constraints to local neighborhoods, which reduces interference with the global semantic structure of the pre-trained VLM.
Second, we introduce neural-collapse-driven discriminability optimization with three losses that together shape the intra-cluster geometry for better inter-class separation and more stable intra-class alignment.
(1) Textual ETF separation loss enforces an ETF constraint among intra-cluster textual features to improve local inter-class separation.
(2) Class-wise convergence loss pulls visual features toward their corresponding textual feature, minimizing intra-class angular deviation.
(3) Rotation stabilization loss anchors each textual feature to its corresponding frozen VLM feature by minimizing their absolute distance to enhance training stability.
Overall, our approach offers an efficient solution for long-tailed recognition when adapting pre-trained VLMs to class-imbalanced data.
We evaluate CPT on base-to-new, cross-dataset, and domain generalization across 11 diverse datasets, and CPT consistently outperforms state-of-the-art methods.
In summary, our contributions are as follows:
\begin{itemize}
\item We propose CPT, which introduces cluster-aware neural collapse constraints, enhancing tail-class discriminability while preserving the generalizability of pre-trained VLMs.
\item We propose three losses: textual ETF separation loss, class-wise convergence loss, and rotation stabilization loss, which together constrain the intra-cluster embedding space for better inter-class separation and intra-class alignment.
\item We conduct extensive experiments on 11 diverse datasets to validate CPT, demonstrating superior tail-class discriminability and generalization.
\end{itemize}
\section{Related Work}
\noindent \textbf{Vision-Language Models:}
Vision-Language models\cite{zhai2022lit, yao2021filip, yuan2021florence} have emerged as transformative tools for multimodal understanding, integrating visual and textual modalities to encode powerful and transferable representations. Early models such as CLIP~\cite{Radford2021CLIP} and ALIGN~\cite{jia2021scaling} use large-scale image-text pairs and contrastive objectives to build strong joint embeddings, showing strong generalization in zero-shot classification, object detection, and cross-domain transfer. Recent work further extends this line toward finer-grained and broader multimodal modeling, including change captioning, active learning, open-set sample selection, fine-grained multimodal LLMs, and audio-visual dubbing~\cite{tu2024smart,zhang2024inductive,tang2026lmda,peng2026fgmllm_survey,li2025dubbing}.
\noindent \textbf{Prompt tuning: }
Prompt tuning \cite{ma2023understandingmitigatingoverfittingprompt, CoPrompt, Prompt_KD} has become a powerful paradigm for adapting pre-trained vision-language models, such as CLIP and ALIGN, to a variety of downstream tasks. By introducing task-specific prompts, it enables efficient adaptation with minimal parameter updates, while enhancing the generalizability of VLMs across different domains and categories. This approach has been successfully applied to tasks including few-shot image recognition \cite{gao2021clip, zhang2021tip, zhou2022conditional, Ye2022KGUCH}, object detection \cite{rasheed2022bridging, cui2025debiased, cui2024stochastic}, and beyond.
\begin{figure*}[!ht]
\centering
{\includegraphics[width=1\textwidth]{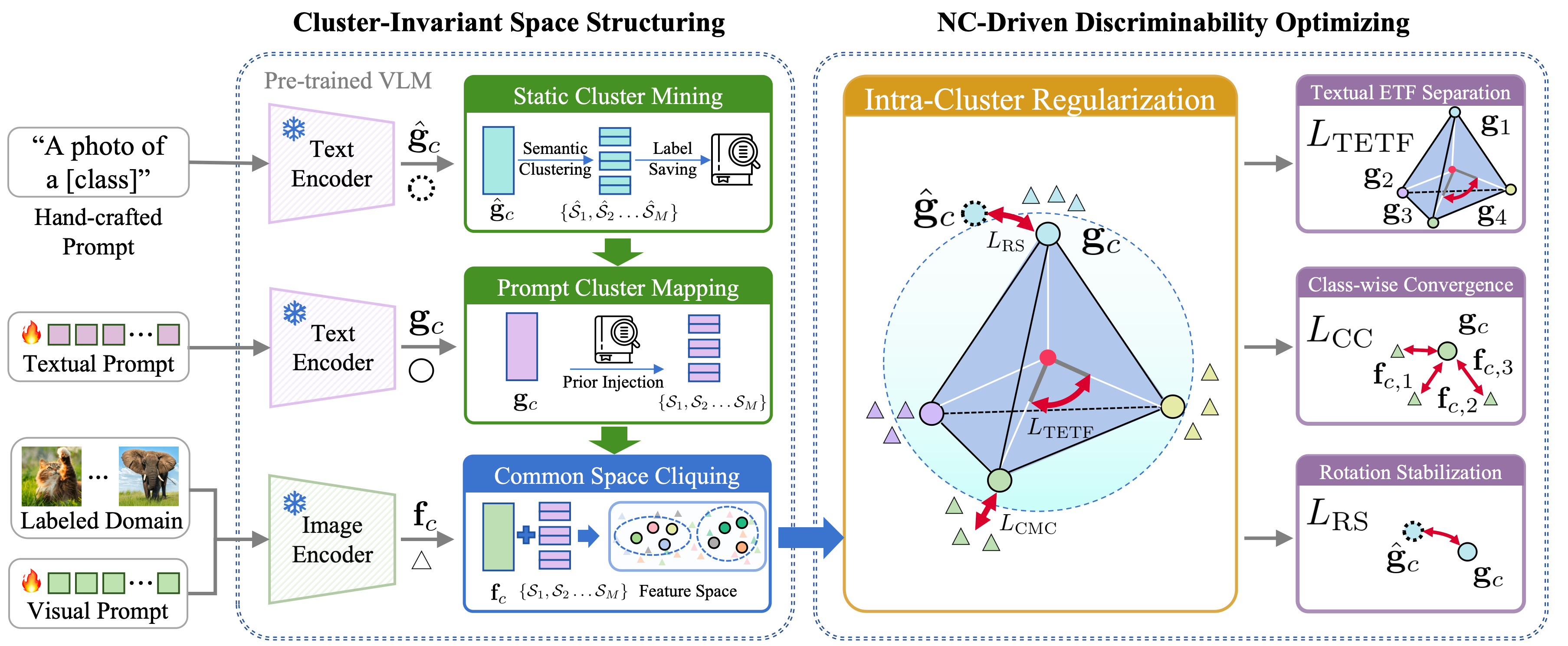}}\vspace{-0.5em}
\caption{
Overview of our proposed cluster-aware neural collapse prompt tuning (CPT). CPT introduces cluster-invariant space structuring to partition pre-trained features into stable semantic clusters, preserving global semantic flexibility.
Within each cluster, it employs NC-driven discriminability optimization  with three targeted loss functions, enhancing local separability and training stability.
}
\label{fig:main_figure}
\end{figure*}
\noindent \textbf{Neural Collapse: }
Neural Collapse (NC) describes a geometric regime that often emerges at the end of supervised training: (i) features from the same class concentrate around a single prototype (“class mean”), (ii) different class prototypes become maximally separated and approximately form a simplex Equiangular Tight Frame (ETF) with uniform pairwise angles, and (iii) the classifier weights align with these prototypes \cite{Papyan_2020,Fang2021MinorityCollapse, Ji2021UnconstrainedLayerPeeled, TirerBruna2022ExtendedUFM, E2021SimplexSymmetry, Zhu2021GeometricAnalysisNC}.
This structure is attractive for long-tailed recognition because it increases inter-class margins and can help prevent tail classes from being absorbed by dominant head classes. Motivated by this, recent prompt tuning methods for vision-language models attempt to directly impose an ETF-like arrangement on all class textual prototypes. In these approaches, the textual prompts of every class are optimized so that the resulting prototypes become nearly equiangular, effectively enforcing global, uniform separation across the entire label space \cite{didi2024NCprompt, sedov2024exploringembeddingpriorsprompttuning}.
\noindent \textbf{Long-Tailed Adaptation and Logit Adjustment:}
Long-tailed recognition has traditionally been addressed by loss reweighting, class-balanced sampling, or logit adjustment based on estimated class priors \cite{Cui2019ClassBalanced,Cao2019LDAM,Ren2020BalancedSoftmax,Menon2021LogitAdjust}. These techniques explicitly favor under-represented (tail) classes by either amplifying their contribution to the loss or directly shifting the decision boundary in their favor \cite{Cao2019LDAM,Menon2021LogitAdjust}. Recent efforts have begun to bring this idea into the vision-language setting by applying class-frequency-aware calibration or logit-adjusted training during prompt tuning \cite{Dong2023LPT,Zhu2023GLA,Hou2025CAPT}. These approaches are attractive because they are simple, require minimal architectural change, and often improve recall on rare classes in-distribution \cite{Menon2021LogitAdjust,Dong2023LPT,Zhu2023GLA}.
Despite these advantages, such reweighting-style or logit-adjusted methods operate primarily at the classifier level: they change how scores are produced, not how features are organized \cite{Kang2020Decoupling,Ren2020BalancedSoftmax,Menon2021LogitAdjust}. They do not restructure the joint image–text embedding space that the vision–language model relies on for transfer. In practice, this can introduce a new failure mode. While head–tail balance improves on the source task, the modified decision surface may deviate from the semantic geometry learned during large-scale pre-training, and full fine-tuning or naive rebalancing can hurt generalization and robustness under domain shift \cite{Dong2023LPT}. In other words, reweighting compensates for class frequency but does not explicitly recover a stable, semantically meaningful arrangement of prototypes in feature space \cite{Kang2020Decoupling}.
Our approach is designed to address exactly this gap. Instead of only shifting logits, we directly shape the representation space in a way that (i) improves tail-class separability and (ii) preserves the high-level semantic structure of the pre-trained model.
\section{Methodology}
\label{sec: Methodology}
Our goal is to adapt a pre-trained vision-language model such as CLIP to class-imbalanced downstream tasks in a way that (i) improves separability for tail classes, (ii) preserves the global semantic structure that enables transfer, and (iii) remains stable across random seeds.
We first review the basic setup of prompt tuning. We then present our two core components.
\subsection{Preliminaries}
\label{sec:3.1}
\textbf{Prompt-tuned CLIP.}
We adopt a CLIP-style model with a frozen image encoder $G_v$ and a frozen text encoder $G_t$. Instead of updating all weights, we insert learnable continuous prompts into both encoders. For an image $x$, the visual encoder with prompts produces an image feature $\mathbf{f} \in \mathbb{R}^d$. For a class label $c$, the text encoder with prompts produces a class textual feature $\mathbf{g}_c \in \mathbb{R}^d$. The original pre-trained textual feature for the same class without learnable prompts is denoted $\hat{\mathbf{g}}_c$. CLIP is trained with a contrastive objective that aligns matched image–text pairs and separates mismatched pairs:
\begin{equation}
\mathcal{L}_{\text{CLIP}} = - \sum_{i=1}^{B} \log
\frac{\exp(\mathrm{sim}(\mathbf{g}_i, \mathbf{f}_i)/\tau)}
{\sum_{j=1}^{B} \exp(\mathrm{sim}(\mathbf{g}_i, \mathbf{f}_j)/\tau)},
\end{equation}
where $\mathrm{sim}(\cdot,\cdot)$ is cosine similarity and $\tau$ is a temperature.
We operate in the prompt tuning regime: only the prompts are updated, while the backbone parameters remain frozen. This is attractive for efficiency and transferability, but on class-imbalanced data it tends to overfit head classes and underfit tail classes.
\subsection{Cluster-Invariant Space Structuring}
The purpose of cluster-invariant space structuring is to (i) preserve the high-level semantic assignments of the pre-trained VLM, which is crucial for transfer and OOD generalization, and (ii) create local semantic fences where we can safely impose stronger geometric constraints within local neighborhoods. To achieve this, we perform the following three steps:
\textbf{Step 1: Static Cluster Mining. }
For each class $c\in\mathcal{C}$, we first extract its textual feature $\hat{\mathbf{g}}_c$ from the VLM text encoder. We run K-means once on $\{\hat{\mathbf{g}}_c\}_{c=1}^K$ to obtain $M$ disjoint semantic clusters $\{\hat{\mathcal{S}}_1, \dots, \hat{\mathcal{S}}_M\}$. Let $\hat{\mathcal{S}}_m$ denote the set of classes assigned to cluster $m$, and let $\hat{\boldsymbol{\mu}}_m$ denote its centroid:
\begin{equation}
\hat{\boldsymbol{\mu}}_m = \frac{1}{|\hat{\mathcal{S}}_m|} \sum_{c \in \hat{\mathcal{S}}_m} \hat{\mathbf{g}}_c .
\end{equation}
We do not re-cluster during training for two reasons. First, the frozen partition preserves the high-level semantic structure learned during pre-training, which plays a crucial role in preserving the generalizability of the model. Second, allowing cluster membership to change with every gradient step would cause the cluster-level objectives to receive highly unstable targets, injecting noise into the optimization process and impeding convergence.
\textbf{Step 2: Prompt Cluster Mapping. }
To leverage the frozen clusters $\{\hat{\mathcal{S}}_1, \dots, \hat{\mathcal{S}}_M\}$ constructed in the previous step, we define a fixed mapping from classes to cluster indices:
\begin{equation}
\pi: \mathcal{C} \rightarrow \{1, \dots, M\}, \quad \pi(c) = m \Longleftrightarrow c \in \hat{\mathcal{S}}_m.
\end{equation}
This mapping $\pi$ encodes the class-to-cluster assignments and acts as a conduit to transfer the semantic structure preserved during pre-training into the training phase. Using $\pi$, we define a set of training-time clusters $\{\mathcal{S}_1, \dots, \mathcal{S}_M\}$ as follows:
\begin{equation}
\mathcal{S}_m = \{ c \in \mathcal{C} \mid \pi(c) = m \}, \quad \forall m \in \{1, \dots, M\}.
\end{equation}
\textbf{Step 3: Common Space Cliquing. }
Each $\mathcal{S}_m$ thus inherits its class composition directly from the corresponding frozen cluster $\hat{\mathcal{S}}_m$. All learnable textual features $\mathbf{g}_c$ and their associated visual features  $\{\mathbf{f}_{c,i}\}_{i=1}^{N_c}$ are consistently aligned with their assigned semantic cluster:
\begin{equation}
\left\{ \mathbf{g}_c \right\} \cup \left\{ \mathbf{f}_{c,i} \right\}_{i=1}^{N_c} \subseteq \mathcal{S}_{\pi(c)}.
\end{equation}
The invariant clustering structure provides a stable semantic structure. We allow different clusters to remain separated according to the pre-trained geometry. This preserves high-level relationships and maintains a richer, higher-rank similarity structure.
\subsection{NC-Driven Discriminability Optimization}
Given the semantic structure above, we impose Neural Collapse–inspired constraints within each semantic cluster. This improves separability among classes, while not destroying the global structure across clusters. We introduce three complementary losses.
\textbf{(1) Textual ETF Separation Loss ($\mathcal{L}_{\text{TETF}}$).}
Due to the design of CLIP, each class has only one text embedding generated from its label. We use this embedding as the prototype of the class.
Within each cluster $S_m$, we collect textual prototypes of all classes:
\begin{equation}
P_m = [\, \mathbf{g}_{c_1}, \mathbf{g}_{c_2}, \dots, \mathbf{g}_{c_{k_m}} \,] \in \mathbb{R}^{d \times k_m}, \quad c_i \in S_m,
\end{equation}
and normalize them:
\begin{equation}
\tilde{P}_m =
\left[
\frac{\mathbf{g}_{c_1}}{\|\mathbf{g}_{c_1}\|_2}, \dots,
\frac{\mathbf{g}_{c_{k_m}}}{\|\mathbf{g}_{c_{k_m}}\|_2}
\right].
\end{equation}
We then form the cosine-similarity matrix
\begin{equation}
C_m = \tilde{P}_m^\top \tilde{P}_m \in \mathbb{R}^{k_m \times k_m}.
\end{equation}
An ideal ETF inside the cluster would produce uniform off-diagonal similarity (equal pairwise angles). We encourage this via
\begin{equation}
\mathcal{L}_{\text{TETF}} =
\frac{1}{M} \sum_{m=1}^{M}
\left\|
C_m + \frac{1}{k_m-1}(\mathbf{1}-I)
\right\|_F^2 ,
\end{equation}
where $\mathbf{1}$ is the all-ones matrix and $I$ is the identity. Intuitively, $\mathcal{L}_{\text{TETF}}$ increases angular separation among classes that occupy the same semantic neighborhood, which is exactly where tail classes are most likely to be confused with dominant head classes.
Unlike a global ETF, $\mathcal{L}_{\text{TETF}}$ never forces unrelated clusters to become equidistant. Thus we gain local discriminability (especially for rare classes) without globally flattening the semantic hierarchy.
\textbf{(2) Class-wise Convergence Loss ($\mathcal{L}_{\text{CC}}$).}
Neural Collapse predicts that, at convergence, all samples from a class align with a single class prototype. We mimic that behavior across modalities by making visual features collapse toward their corresponding textual prototype. For class $c$ with $N_c$ training samples $\{\mathbf{f}_{c,i}\}_{i=1}^{N_c}$, we define:
\begin{equation}
\mathcal{L}_{\text{CC}} =
\frac{1}{K}
\sum_{c=1}^{K}
\left(
\frac{1}{N_c}
\sum_{i=1}^{N_c}
\left\|
\frac{\mathbf{f}_{c,i}}{\|\mathbf{f}_{c,i}\|_2} -
\frac{\mathbf{g}_c}{\|\mathbf{g}_c\|_2}
\right\|_2^2
\right).
\end{equation}
This loss reduces intra-class angular spread and enforces cross-modal alignment: the image features for class $c$ are explicitly pulled toward the prompt-tuned textual prototype $\mathbf{g}_c$. As a result, the textual prototype acts as a cluster-anchored semantic center for that class.
\textbf{(3) Rotation Stabilization Loss ($\mathcal{L}_{\text{RS}}$).}
ETF-style constraints specify relative geometry (pairwise angles) but leave the absolute orientation of the prototypes underdetermined. Any global rotation of all prototypes preserves pairwise angles and thus leaves $\mathcal{L}_{\text{TETF}}$ nearly unchanged.
Because CLIP-style models rely on consistent alignment between the vision and text encoders, this uncontrolled drift manifests as higher seed-to-seed variance in downstream accuracy.
We remove this source of instability by softly anchoring each learnable textual prototype $\mathbf{g}_c$ to its frozen pre-trained prototype $\hat{\mathbf{g}}_c$. We define:
\begin{equation}
\mathcal{L}_{\text{RS}} =
\frac{1}{K} \sum_{c=1}^{K}
\left\| \mathbf{g}_c - \hat{\mathbf{g}}_c \right\|_1 .
\end{equation}
This term penalizes large absolute deviations from the original pre-trained representation while still allowing moderate adaptation. Geometrically, $\mathcal{L}_{\text{RS}}$ fixes the otherwise free global rotation and prevents prototype drift across seeds. Empirically, this produces substantially lower variance across random seeds and more reproducible performance.
\begin{table}[t]
  \centering
  \footnotesize
  \setlength{\tabcolsep}{3pt}
  \renewcommand{\arraystretch}{1.08}
\begin{tabular}{lcccc|ccc}
    \toprule
    \multicolumn{8}{c}{\textbf{Imbalance Ratio $\tau = 0.25$}}\\
    \midrule
    \multicolumn{4}{c}{\textbf{Components}} & & \multicolumn{3}{c}{\textbf{Results}}\\
    \cmidrule(r){1-4}\cmidrule(l){6-8}
    \textbf{} & \textbf{$\mathcal{L}_{\text{TETF}}$} & \textbf{$\mathcal{L}_{\text{CC}}$} & \textbf{$\mathcal{L}_{\text{RS}}$} &&
    \textbf{Base-to-New} & \textbf{Cross-Dataset} & \textbf{Domain Gen.}\\
    \midrule
    (1)                              &        &        &        && 70.35 & 63.89 & 57.99 \\
    (2)                              & \cmark &        &        && 72.03 & 65.75 & 59.21 \\
    (3)                              & \cmark & \cmark &        && 73.22 & 66.03 & 59.66 \\
    (4)                              & \cmark &        & \cmark && 72.42  & 65.94 & 59.49 \\
    \rowcolor{gray!20}
    (5)                              & \cmark & \cmark & \cmark && \textbf{73.58} & \textbf{66.76} & \textbf{60.16} \\
    \bottomrule
  \end{tabular}
  \vspace{0.6em}
 \begin{tabular}{lcccc|ccc}
    \multicolumn{8}{c}{\textbf{Imbalance Ratio $\tau = 0.06$}}\\
    \midrule
    \multicolumn{4}{c}{\textbf{Components}} & & \multicolumn{3}{c}{\textbf{Results}}\\
    \cmidrule(r){1-4}\cmidrule(l){6-8}
    \textbf{} & \textbf{$\mathcal{L}_{\text{TETF}}$} & \textbf{$\mathcal{L}_{\text{CC}}$} & \textbf{$\mathcal{L}_{\text{RS}}$} &&
    \textbf{Base-to-New} & \textbf{Cross-Dataset} & \textbf{Domain Gen.}\\
    \midrule
    (1)                              &        &        &        && 69.99 & 63.84 & 57.03 \\
    (2)                              & \cmark &        &        && 71.10 & 64.88 & 57.92 \\
    (3)                              & \cmark & \cmark &        && 71.85 & 65.49 & 58.56 \\
    (4)                              & \cmark &        & \cmark && 71.93  & 65.63 & 58.64  \\
    \rowcolor{gray!20}
    (5)                              & \cmark & \cmark & \cmark && \textbf{72.47} & \textbf{65.97} & \textbf{59.12} \\
    \bottomrule
  \end{tabular}
  \vspace{0.35em}
  \footnotesize
  \caption{Ablation study of components in CPT under two imbalance ratios. Results are reported as harmonic mean over 11 datasets.}
  \label{tab:ablations_components}
\end{table}

\begin{figure}[t]
\centering
\includegraphics[width=1\linewidth]{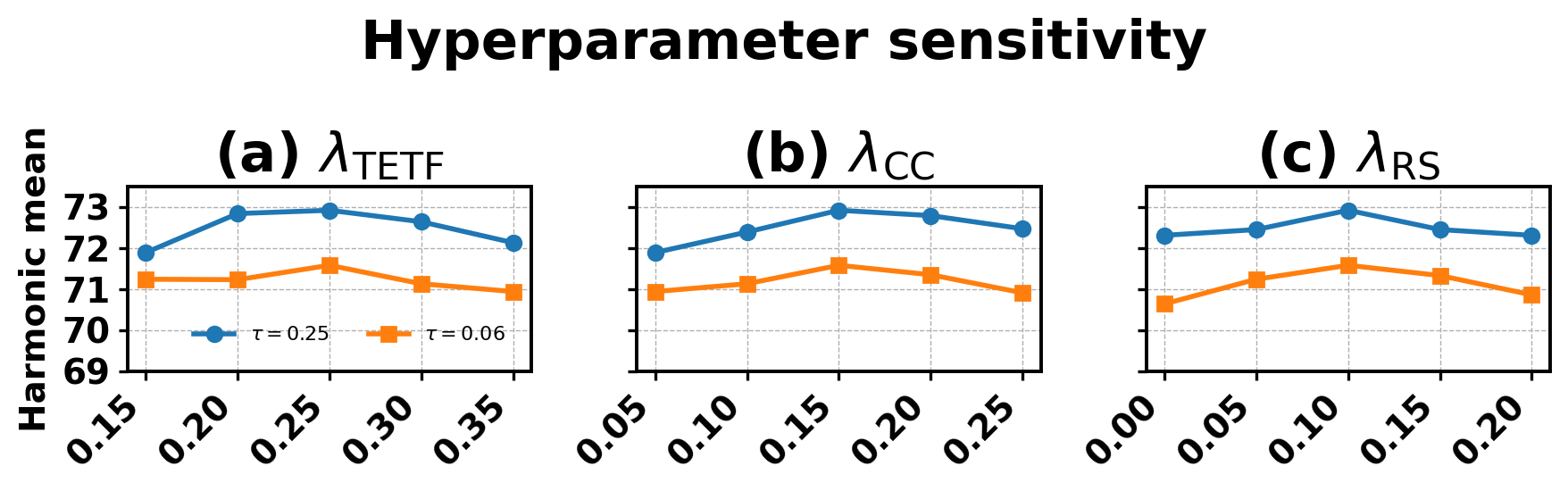}
\caption{Sensitivity of CPT to the loss weights $\lambda_{\text{TETF}}$, $\lambda_{\text{CC}}$, and $\lambda_{\text{RS}}$ on the ImageNet base-to-new setting. We vary one coefficient while fixing the other two at their default values: $\lambda_{\text{TETF}} = 0.25, \lambda_{\text{CC}} = 0.15, \lambda_{\text{RS}} = 0.10.$
We plot harmonic mean accuracy under two imbalance levels ($\tau { = } 0.25$ and $\tau {=} 0.06$).
}
\label{fig:ablation:loss}
\vspace{-1.0em}
\end{figure}
\subsection{Overall Objective}
The final training objective is the sum of the CLIP contrastive loss and the three cluster-aware Neural Collapse losses:
\begin{equation}
\mathcal{L}_{\text{total}} =
\mathcal{L}_{\text{CLIP}}
+ \lambda_{\text{TETF}} \mathcal{L}_{\text{TETF}}
+ \lambda_{\text{CC}}   \mathcal{L}_{\text{CC}}
+ \lambda_{\text{RS}}   \mathcal{L}_{\text{RS}} .
\end{equation}
We report a sensitivity analysis of these hyperparameters in Section 4.5.
In summary, CPT helps preserve the high-level semantic structure of the pre-trained model, enforces stronger separation within local neighborhoods, aligns visual features with textual prototypes, and anchors prototype orientations to stabilize optimization.
As a result, CPT improves tail-class separability and generalization under class imbalance.
\begin{table*}[t]
  \centering
  \footnotesize
  \setlength{\tabcolsep}{8pt}
  \renewcommand{\arraystretch}{1.08}
\begin{tabular}{c|c c c c c c c c c c c | >{\columncolor{gray!20}}c}
    \toprule
    \multicolumn{13}{c}{\textbf{Imbalance Ratio $\tau = 1$ (Balance) }}\\
    \midrule
    Method & IN.  & Cal. & OP. & SC. & Flw. & Food. & FA. & SUN. & DTD & ES. & UCF. & Avg\\
    \midrule
    MaPLe(CVPR'23) & 73.47 & 96.02 & 96.28 & 73.47 & 82.56 & 91.38 & 36.50 & 79.75 & 68.16 & 82.35 & 80.77 & 78.25 \\
    CoPrompt(ICLR'24) & 72.86 & 95.94 & 96.33 & 75.47 & 84.51 & 90.95 & 37.82 & 79.29 & 68.93 & 83.07 & 80.78 & 78.72 \\
    NPT(KDD'24) & 73.82 & 96.23 & 96.24 & 72.49 & 82.37 & 91.67 & 36.46 & 79.76 & 68.88 & 82.12 & 81.42 & 78.31 \\
    DeKg(ICLR'25) & 73.07 & 96.89 & 96.09 & 77.83 & 84.51 & 91.14 & 39.51 & 80.35 & 69.70 & 87.42 & 84.80 & \underline{80.12} \\
    DPC(CVPR'25) & 73.85 & 96.50 & 96.71 & 78.46 & 83.59 & 91.49 & 40.24 & 80.76 & 66.84 & 84.13 & 83.21 & 79.62 \\
    \rowcolor{gray!20}
    \textbf{CPT(Ours)} & 73.92 & 96.53 & 96.37 & 77.21 & 84.69 & 91.17 & 40.13 & 80.50 & 71.20 & 87.54 & 83.83 & \textbf{80.28} \\
\bottomrule
\end{tabular}
  \vspace{0.6em}
\begin{tabular}{c|c c c c c c c c c c c | >{\columncolor{gray!20}}c}
\multicolumn{13}{c}{\textbf{Imbalance Ratio $\tau = 0.25$}}\\
\midrule
Method   & IN.  & Cal. & OP. & SC. & Flw. & Food. & FA. & SUN. & DTD & ES. & UCF. & Avg\\
\midrule
MaPLe(CVPR'23)    & 70.93 & 95.38 & 94.22 & 70.25 & 76.59 & 88.71 & 28.62 & 75.98 & 55.15 & 53.97 & 75.43 & 71.38 \\
CoPrompt(ICLR'24) & 71.50 & 94.71 & 95.24 & 70.20 & 77.77 & 89.35 & 29.56 & 76.12 & 56.58 & 57.89 & 75.69 & 72.24 \\
NPT(KDD'24)       & 72.22 & 95.66 & 96.20 & 70.89 & 78.58 & 90.23 & 29.86 & 76.88 & 57.16 & 58.48 & 76.65 & 72.98 \\
DeKg(ICLR'25)     & 72.08 & 95.86 & 95.30 & 71.12 & 78.98 & 90.46 & 29.91 & 76.90 & 57.58 & 58.61 & 76.11 & 72.99 \\
DPC(CVPR'25)      & 71.85 & 95.35 & 96.26 & 71.41 & 78.11 & 90.00 & 29.74 & 77.18 & 57.08 & 58.85 & 76.28 & 72.92 \\
\rowcolor{gray!20}
\textbf{CPT(Ours)} & 72.62 & 96.44 & 96.16 & 71.58 & 79.37 & 91.13 & 30.16 & 77.65 & 57.74 & 59.09 & 77.42 & \textbf{73.58} \\
\bottomrule
\end{tabular}
  \vspace{0.6em}
\begin{tabular}{c|c c c c c c c c c c c | >{\columncolor{gray!20}}c}
\multicolumn{13}{c}{\textbf{Imbalance Ratio $\tau = 0.06$}}\\
\midrule
Method & IN.  & Cal. & OP. & SC. & Flw. & Food. & FA. & SUN. & DTD & ES. & UCF. & Avg\\
\midrule
MaPLe(CVPR'23)    & 70.10 & 95.53 & 94.21 & 69.29 & 72.29 & 89.74 & 21.89 & 75.00 & 47.37 & 51.10 & 73.85 & 69.12 \\
CoPrompt(ICLR'24) & 70.20 & 94.69 & 93.64 & 69.88 & 73.71 & 89.27 & 27.43 & 75.13 & 56.42 & 56.89 & 74.20 & 71.04 \\
NPT(KDD'24)       & 70.89 & 95.64 & 94.56 & 70.58 & 74.46 & 90.20 & 27.72 & 75.89 & 56.98 & 57.47 & 74.95 & 71.76 \\
DeKg(ICLR'25)     & 70.37 & 95.20 & 94.52 & 70.19 & 73.78 & 90.49 & 27.73 & 75.24 & 56.66 & 57.92 & 74.42 & 71.50 \\
DPC(CVPR'25)      & 70.62 & 95.75 & 94.77 & 70.37 & 74.87 & 90.25 & 27.75 & 75.39 & 56.91 & 57.38 & 75.19 & 71.75 \\
\rowcolor{gray!20}
\textbf{CPT(Ours)} & 71.58 & 96.58 & 95.50 & 71.27 & 75.20 & 91.10 & 27.99 & 76.65 & 57.54 & 58.03 & 75.69 & \textbf{72.47} \\
\bottomrule
\end{tabular}
  \caption{Base-to-New classification accuracy (\%) across 11 downstream datasets under varying class-imbalance ratios. Prompt-based methods are trained from base classes, using ViT-B/16 as the backbone. All reported values in the table indicate the harmonic mean (H) between base and novel class accuracy. The cases with imbalance ratio $\tau = {0.25, 0.06}$ represent highly imbalanced scenarios, designed to assess the robustness of tail-class recognition.}
  \label{tab:base2new}
\end{table*}

\begin{table*}[t]
  \centering
  \footnotesize
  \setlength{\tabcolsep}{8pt}
  \renewcommand{\arraystretch}{1.08}
\begin{tabular}{l|c|cccccccccc|>{\columncolor{gray!20}}c}
    \toprule
    \multicolumn{13}{c}{\textbf{Imbalance Ratio $\tau = 1$ (Balance) }}\\
    \midrule
    Method & IN.   & Cal. & OP. & SC. & Flw. & Food. & FA. & SUN. & DTD & ES. & UCF. & Avg\\
    \midrule
    MaPLe(CVPR'23)      & 70.72 & 93.53 & 90.49 & 65.57 & 72.23 & 86.20 & 24.74 & 67.01 & 46.49 & 48.06 & 68.69 & 66.30 \\
    CoPrompt(ICLR'24)   & 70.10 & 94.40 & 91.13 & 66.67 & 72.60 & 86.23 & 25.40 & 68.92 & 47.73 & 49.77 & 69.43 & \underline{67.23} \\
    NPT(KDD'24)         & 70.70 & 94.50 & 90.64 & 66.73 & 72.62 & 86.21 & 25.23 & 68.37 & 47.63 & 50.25 & 68.87 & 67.11 \\
    DeKg(ICLR'25)       & 72.33 & 94.73 & 90.68 & 66.05 & 73.20 & 86.59 & 25.05 & 67.19 & 45.61 & 51.37 & 68.78 & 66.92 \\
    DPC(CVPR'25)        & 71.42 & 93.60 & 90.25 & 65.70 & 71.73 & 86.15 & 23.90 & 67.10 & 46.87 & 48.68 & 68.75 & 66.27 \\
    \rowcolor{gray!20}
    \textbf{CPT(Ours)}  & 71.40 & 95.42 & 91.54 & 67.38 & 73.35 & 87.09 & 25.49 & 69.03 & 48.13 & 50.74 & 69.53 & \textbf{67.78} \\
    \bottomrule
\end{tabular}
\vspace{0.6em}
\begin{tabular}{l|c|cccccccccc|>{\columncolor{gray!20}}c}
    \multicolumn{13}{c}{\textbf{Imbalance Ratio $\tau = 0.25$ }}\\
    \midrule
    Method & IN.   & Cal. & OP. & SC. & Flw. & Food. & FA. & SUN. & DTD & ES. & UCF. & Avg\\
    \midrule
    MaPLe(CVPR'23)      & 69.20 & 92.63 & 90.34 & 65.17 & 71.54 & 85.03 & 21.39 & 65.57 & 45.84 & 47.83 & 65.34 & 65.07 \\
    CoPrompt(ICLR'24)   & 69.34 & 92.64 & 89.76 & 65.27 & 71.89 & 85.35 & 22.54 & 66.24 & 45.58 & 49.20 & 66.42 & 65.49 \\
    NPT(KDD'24)         & 69.40 & 93.60 & 90.40 & 65.95 & 72.33 & 86.19 & 22.78 & 66.89 & 46.05 & 49.67 & 67.12 & \underline{66.10} \\
    DeKg(ICLR'25)       & 69.46 & 93.79 & 90.02 & 65.49 & 72.39 & 86.59 & 22.60 & 66.96 & 44.57 & 49.96 & 67.33 & 65.97 \\
    DPC(CVPR'25)        & 69.21 & 92.63 & 89.96 & 64.65 & 70.36 & 85.33 & 21.44 & 65.75 & 45.60 & 45.50 & 65.76 & 64.70 \\
    \rowcolor{gray!20}
    \textbf{CPT(Ours)}  & 69.89 & 94.56 & 91.27 & 66.64 & 73.09 & 87.03 & 23.02 & 67.54 & 46.53 & 50.16 & 67.82 & \textbf{66.76} \\
    \bottomrule
\end{tabular}
\vspace{0.6em}
\begin{tabular}{l|c|cccccccccc|>{\columncolor{gray!20}}c}
    \multicolumn{13}{c}{\textbf{Imbalance Ratio $\tau = 0.06$ }}\\
    \midrule
    Method & IN.   & Cal. & OP. & SC. & Flw. & Food. & FA. & SUN. & DTD & ES. & UCF. & Avg\\
    \midrule
    MaPLe(CVPR'23)      & 68.32 & 92.23 & 90.07 & 64.38 & 70.37 & 84.77 & 20.47 & 64.53 & 45.17 & 47.03 & 64.40 & 64.34 \\
    CoPrompt(ICLR'24)   & 68.22 & 92.35 & 89.01 & 64.64 & 71.02 & 85.35 & 21.29 & 65.38 & 44.54 & 47.90 & 65.53 & 64.70 \\
    NPT(KDD'24)         & 68.90 & 93.31 & 89.93 & 65.28 & 71.77 & 86.00 & 21.51 & 66.07 & 44.97 & 48.17 & 66.20 & 65.32 \\
    DeKg(ICLR'25)       & 69.42 & 92.89 & 89.63 & 65.07 & 72.02 & 85.99 & 21.35 & 66.17 & 44.47 & 49.90 & 65.86 & \underline{65.33} \\
    DPC(CVPR'25)        & 68.23 & 91.79 & 89.09 & 64.40 & 70.25 & 84.68 & 21.14 & 65.35 & 44.17 & 44.20 & 64.46 & 63.95 \\
    \rowcolor{gray!20}
    \textbf{CPT(Ours)}  & 69.58 & 94.23 & 90.82 & 65.92 & 72.50 & 86.86 & 21.71 & 66.71 & 45.42 & 48.65 & 66.89 & \textbf{65.97} \\
    \bottomrule
\end{tabular}
  \caption{Cross-dataset generalization accuracy (\%) under varying class-imbalance ratios. All other settings follow the Base-to-New classification.}
  \label{tab:cross_dataset}
\end{table*}

\begin{table}[t]
  \centering
  \footnotesize
  \setlength{\tabcolsep}{6pt}
  \renewcommand{\arraystretch}{1.08}
\begin{tabular}{l|c|cccc|>{\columncolor{gray!20}}c}
\toprule
\multicolumn{7}{c}{\textbf{Imbalance Ratio $\tau = 1$ (Balance) }}\\
\midrule
Method & IN. & IN.-V2 & IN.-S & IN.-A & IN.-R & Avg\\
\midrule
CoCoOp     & 71.02 & 64.07 & 48.75 & 50.63 & 76.18 & 59.91 \\
MaPLe      & 70.72 & 64.07 & 49.15 & 50.90 & 76.98 & 60.27 \\
NPT        & 70.70 & 64.23 & 48.10 & 49.70 & 76.89 & 59.73 \\
CoPrompt   & 70.10 & 64.15 & 48.83 & 50.58 & 77.31 & 60.22 \\
DeKg       & 72.33 & 64.21 & 48.82 & 50.62 & 76.64 & 60.07 \\
DPC        & 71.42 & 64.05 & 49.55 & 50.65 & 77.50 & \underline{60.44} \\
\rowcolor{gray!20}
\textbf{CPT(Ours)} & 71.40 & 64.23 & 49.40 & 50.66 & 77.70 & \textbf{60.50} \\
\bottomrule
\end{tabular}
\vspace{0.6em}
\begin{tabular}{l|c|cccc|>{\columncolor{gray!20}}c}
\multicolumn{7}{c}{\textbf{Imbalance Ratio $\tau = 0.25$ }}\\
\midrule
Method & IN. & IN.-V2 & IN.-S & IN.-A & IN.-R & Avg\\
\midrule
CoCoOp     & 68.43 & 62.21 & 46.89 & 48.56 & 75.64 & 58.33 \\
MaPLe      & 69.20 & 62.40 & 47.70 & 48.98 & 75.83 & 58.73 \\
NPT        & 69.40 & 63.10 & 47.83 & 49.10 & 76.20 & 59.06 \\
CoPrompt   & 69.34 & 62.47 & 48.37 & 49.60 & 76.46 & 59.22 \\
DeKg       & 69.79 & 63.34 & 48.72 & 50.51 & 76.37 & 59.73 \\
DPC        & 69.71 & 63.12 & 49.25 & 50.64 & 76.85 & \underline{59.96} \\
\rowcolor{gray!20}
\textbf{CPT(Ours)} & 69.89 & 63.74 & 49.35 & 50.64 & 76.89 & \textbf{60.16} \\
\bottomrule
\end{tabular}
\vspace{0.6em}
\begin{tabular}{l|c|cccc|>{\columncolor{gray!20}}c}
\multicolumn{7}{c}{\textbf{Imbalance Ratio $\tau = 0.06$ }}\\
\midrule
Method & IN. & IN.-V2 & IN.-S & IN.-A & IN.-R & Avg\\
\midrule
CoCoOp     & 66.21 & 60.35 & 46.13 & 48.25 & 74.56 & 57.32 \\
MaPLe      & 67.32 & 62.33 & 47.20 & 48.47 & 74.47 & 58.12 \\
NPT        & 68.70 & 62.83 & 47.53 & 48.07 & 74.70 & 58.28 \\
CoPrompt   & 68.22 & 62.20 & 48.03 & 48.56 & 74.92 & 58.43 \\
DeKg       & 69.20 & 62.64 & 48.38 & 48.56 & 75.12 & \underline{58.68} \\
DPC        & 69.19 & 62.77 & 49.00 & 48.56 & 75.27 & 58.90 \\
\rowcolor{gray!20}
\textbf{CPT(Ours)} & 69.58 & 63.43 & 49.02 & 48.56 & 75.46 & \textbf{59.12} \\
\bottomrule
\end{tabular}
  \caption{Domain generalization accuracy (\%) under varying class-imbalance ratios. }
  \label{tab:domain_gen}
\end{table}

\section{Experiments}
\subsection{Evaluation settings}
Following DPC~\cite{li2025dpc}, DeKg~\cite{Li_2025_ICLR} and NPT~\cite{didi2024NCprompt}, we evaluate our CPT under base-to-new generalization, domain generalization, and cross-dataset transfer generalization over 11 image classification benchmark datasets~\cite{imagenet,caltech101,oxford_pets,stanford_cars,flowers102,food101,aircraft,eurosat,ucf101,dtd,sun397}.
To simulate imbalance, we downsample each class to follow an exponential decay distribution, controlled by imbalance ratios $\tau \in \{1, 0.25, 0.06\}$. Here, $\tau$ is defined as the ratio between the smallest and largest class sizes, i.e., $\tau = \min\{n_k\} / \max\{n_k\}$, where $n_k$ denotes the number of training samples in the $k$-th class. We fix $\max\{n_k\} = 16$ across all settings.
Refer to the supplementary material for implementation details.
\subsection{Base-to-new Generalization}
We follow a base-to-new evaluation protocol across 11 diverse classification datasets. Each dataset is evenly split into base and novel classes. To thoroughly evaluate the effectiveness of CPT in terms of generalizability and tail-class discriminability, we conduct experiments under three different imbalance settings, corresponding to imbalance ratios $\tau \in \{1, 0.25, 0.06\}$. $\tau$ is defined as the ratio between the smallest and largest class sizes, i.e., $\tau = \min\{n_k\} / \max\{n_k\}$, where $n_k$ denotes the number of training samples in the $k$-th class. We fix $\max\{n_k\} = 16$ across all settings.
As shown in Table~\ref{tab:base2new}, when the data is balanced ($\tau=1$), CPT achieves comparable performance to other leading methods across both base and new classes, indicating that our method does not sacrifice generalization when no imbalance is present.
Under more challenging settings with $\tau=0.25$ and $\tau=0.06$, where long-tailed distributions heavily impact model performance, CPT consistently delivers notable improvements on new classes, demonstrating its ability to enhance tail-class recognition.
These results highlight the key advantage of our design: by preserving the pre-trained semantic structure through cluster-aware constraints, CPT improves tail-class discriminability without compromising generalizability.
\subsection{Cross Dataset Evaluation}
To assess the cross-domain generalizability of CPT, we conduct a cross-dataset transfer experiment, where all models are trained on the full ImageNet-1K dataset and directly evaluated on ten diverse target datasets without further adaptation.
This setup simulates a real-world distribution shift scenario, where the learned features must generalize to novel, out-of-distribution (OOD) domains with varying label spaces and image statistics.
Table~\ref{tab:cross_dataset} reports the performance of different methods across these target datasets.
Compared with existing methods, CPT achieves the highest average accuracy of 67.78\%, outperforming DPC (66.27\%), DeKg (66.92\%), and NPT (67.11\%).
This is consistent with our findings under the settings with $\tau=0.25$ and $\tau=0.06$, where CPT notably enhances tail-class performance across imbalanced domains.
This demonstrates that CPT can preserve the high-level semantic structure learned from pre-training, which enhances generalizability across heterogeneous domains.
\subsection{Domain Generalization Experiments}
To further assess the out-of-distribution robustness of CPT, we conduct domain generalization experiments using the ImageNet dataset as the source domain.
Specifically, we train the model on the original ImageNet-1K dataset and evaluate it on four OOD variants—ImageNet-V2, ImageNet-S (Sketch), ImageNet-A (Adversarial), and ImageNet-R (Rendition).
The results are summarized in Table~\ref{tab:domain_gen}.
Across all four datasets, CPT consistently achieves top-tier performance, surpassing MaPLe, NPT, and CoPrompt.
Even under extreme class imbalance ($\tau=0.06$), CPT exhibits robust tail-class accuracy, confirming the effectiveness of its cluster-aware neural collapse optimization strategy.
\subsection{Ablative Analysis}
\begin{figure}[t]
\centering
\includegraphics[width=1\linewidth]{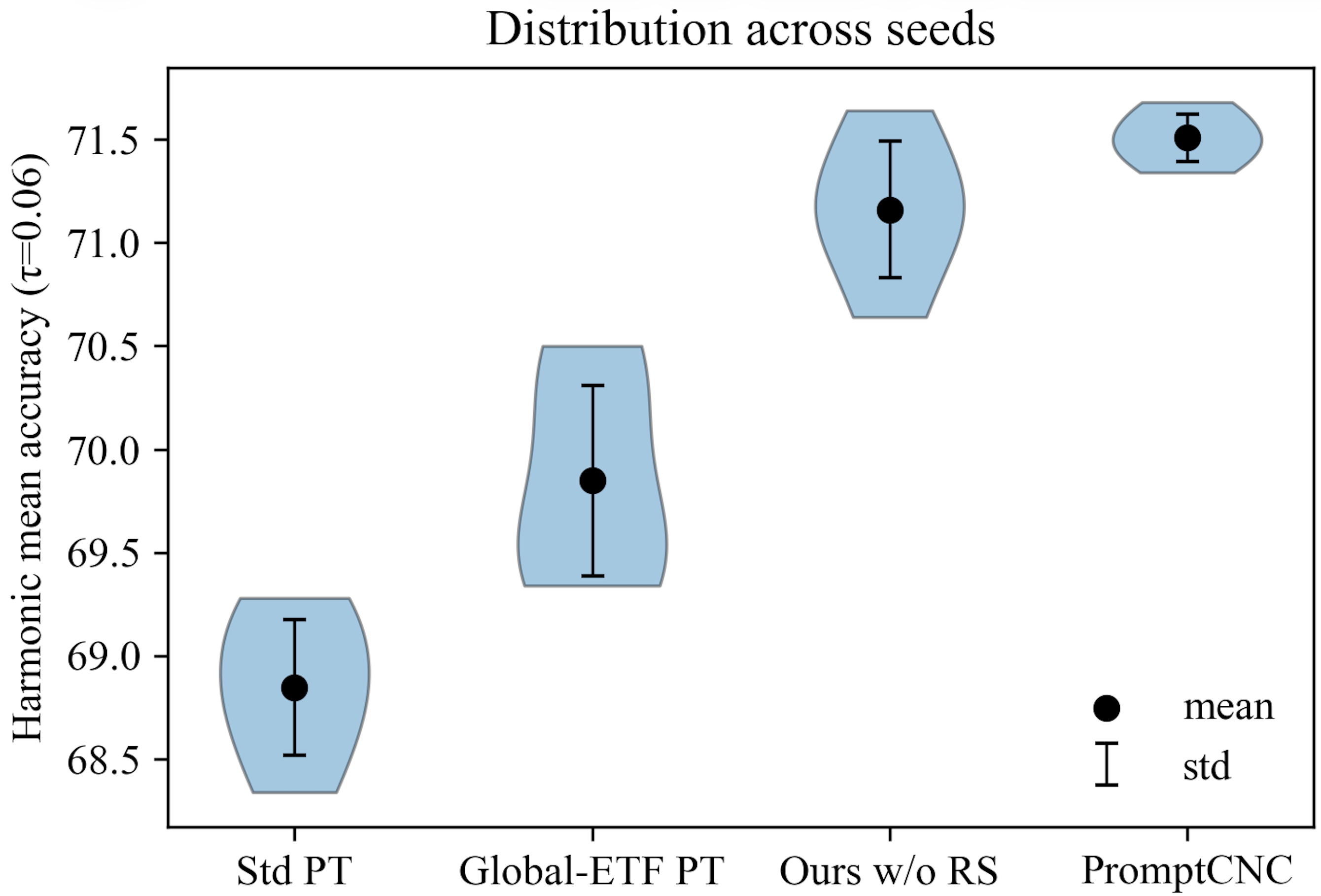}
\caption{
Training stability on class-imbalanced data ($\tau{=}0.06$) in the ImageNet base-to-new setting.
Violin plots illustrate the distribution of harmonic-mean accuracy across five random seeds for four methods, namely Standard PT (COOP), Global-ETF PT, Ours w/o RS, and CPT. The dots indicate the mean, and the error bars represent one standard deviation.
}
\label{fig:ablation:stability}
\vspace{-1.0em}
\end{figure}

\noindent \textbf{Effectiveness of CPT:}
We conduct an ablation study to assess the contribution of each core component in the CPT framework, as illustrated in Table \ref{tab:ablations_components}.
Starting from a lightweight baseline that only applies instance-level prompt tuning (IVLP), we progressively introduce our three key modules.
Table~\ref{tab:ablations_components} summarizes the performance of each variant, reporting results across three benchmark experiments.
We observe that each loss consistently contributes additional gains on top of the previous configuration.
Under both imbalance ratios, the full model with all three losses achieves the best performance across all metrics.
\noindent \textbf{Hyperparameter Sensitivity:}
To evaluate robustness to the choice of loss weights, we sweep each coefficient around its default value while holding the others fixed, as illustrated in Fig.~\ref{fig:ablation:loss}. Concretely, we vary each hyperparameter around the default setting:
$
\lambda_{\text{TETF}} = 0.25,
\lambda_{\text{CC}}   = 0.15,
\lambda_{\text{RS}}   = 0.10.
$
We report harmonic mean accuracy on the ImageNet base-to-new setting under two imbalance ratios ($\tau = 0.25$ and $\tau = 0.06$). Across these sweeps, performance changes by less than 3\% on average, even in the most imbalanced regime ($\tau = 0.06$).
\noindent \textbf{Training Stability and Seed Variance:}
Prompt tuning is known to be sensitive to random seeds, especially in few-shot and long-tailed regimes.
We quantify stability by repeating each experiment with multiple random seeds and reporting both the mean accuracy and standard deviation, as illustrated in Fig.~\ref{fig:ablation:stability}.
We observe two consistent trends:
Methods that enforce a global ETF without anchoring exhibit large variance across seeds.
This is expected: a global ETF constrains pairwise angles but not absolute orientation.
Different seeds tend to settle into different rotations, leading to run-to-run differences in downstream accuracy.
CPT shows substantially lower standard deviation across seeds, often cutting the variance by roughly half compared to global-ETF baselines.
The rotation stabilization loss $\mathcal{L}_{\text{RS}}$, which anchors each learnable textual prototype $\mathbf{g}_c$ to its frozen pre-trained prototype $\hat{\mathbf{g}}_c$, is critical here.
By penalizing large absolute deviations, $\mathcal{L}_{\text{RS}}$ removes the unconstrained global-rotation degree of freedom and keeps different runs aligned to the same semantic frame of reference.
\begin{table}[t]
\centering
\small
\begin{tabular}{lcccc}
\toprule
\multirow{2}{*}{Clustering} & \multicolumn{2}{c}{$\tau=0.25$} & \multicolumn{2}{c}{$\tau=0.06$} \\
\cmidrule(lr){2-3}\cmidrule(lr){4-5}
& H $\uparrow$ & Std $\downarrow$ & H $\uparrow$ & Std $\downarrow$ \\
\midrule
Cosine K-means      & 72.62 & 0.35 & 71.58 & 0.52 \\
Euclidean K-means   & 72.19 & 0.48 & 71.52 & 0.66 \\
Spectral (cosine)   & 71.72 & 0.74 & 71.39 & 1.10 \\
K-medoids (cosine)  & 71.86 & 0.92 & 71.49 & 1.32 \\
\bottomrule
\end{tabular}
\caption{Clustering choice ablation. We use static clustering on frozen textual prototypes with fixed $M=16$, identical seeds, and cosine-silhouette model selection. We report the base-to-new harmonic mean $H$ for $\tau\in\{0.25,0.06\}$ and the standard deviation across seeds. Cosine-aligned partitioners give higher $H$ and lower variance. Numbers are averaged over datasets; Std is computed across seeds.}
\label{tab:cluster_ablation}
\end{table}

\noindent \textbf{Ablation on Clustering Strategy.}
We design a cluster-invariant space by static clustering on frozen textual features.
We test four algorithms: Euclidean K-means, cosine K-means, spectral (cosine affinity), and K-medoids.
We keep the cluster count $M$ and the seed grid fixed, and we tune only on frozen features with a cosine-silhouette score.
We report the base-to-new harmonic mean $H$ ($\tau\in\{0.25,0.06\}$) and the seed-to-seed standard deviation.
Cosine K-means gives the best $H$ and lower variance than Euclidean K-means; spectral is close but more sensitive; K-medoids is worse in both $H$ and stability.
Table~\ref{tab:cluster_ablation} shows that CPT does not rely on a specific clustering heuristic.
Full analysis and additional ablations, including the effect of the number of samples per cluster, are provided in the supplementary material.
\section{Conclusion}
We address the challenge of enhancing tail-class discriminability in prompt-tuned VLMs without sacrificing their overall generalization.
Specifically, CPT integrates cluster-invariant space structuring that preserves the high-level semantics of the pre-trained VLM, while applying NC-driven discriminability optimization with three losses.
The three targeted losses—Textual ETF Separation, Class-wise Convergence, and Rotation Stabilization—jointly shape intra-cluster geometry for better inter-class separation and stable intra-class alignment.
Extensive experiments on 11 benchmarks show consistent gains under different settings, particularly in long-tailed regimes.
CPT demonstrates stronger tail-class performance than SOTA methods, while maintaining generalization to new classes.
\section*{Acknowledgements}
This work was supported by the National Nature Science Foundation of China (62322211, 62336008), the "Pioneer" and "Leading Goose" R\&D Program of Zhejiang Province(2024C01023).
{
    \small
    \bibliographystyle{ieeenat_fullname}
    \bibliography{egbib}
}
\clearpage
\setcounter{page}{1}
\maketitlesupplementary
\section{Experiments Setting}
\subsection{Training Details.}
We evaluate our CPT under base-to-new generalization, domain generalization, and cross-dataset transfer generalization over 11 image classification benchmark datasets.
We conduct the experiments based on the vision backbone with Vit-B/16.
We set the number of visual and textual prompts to 4, initializing them with the template 'a photo of a []'.
All our models are trained with a batch size of 4.
We utilize SGD as our optimization method, starting with an initial learning rate of 0.0025 and employing a cosine annealing scheduler that includes one warm-up epoch.
Depending on the specific experiment setup, the training duration varies: 1)in the base-to-new setting, models are trained for 12 epochs; 2)in the few-shot setting, the training extends to 25 epochs; 3)in domain generalization and cross-dataset evaluation, we limit training to 8 epochs.
All experiments are conducted based on a single NVIDIA 3090 GPU.
To simulate imbalance, we downsample each class to follow an exponential decay distribution, controlled by imbalance ratios $\tau \in \{1, 0.25, 0.6\}$. Here, $\tau$ is defined as the ratio between the smallest and largest class sizes, i.e., $\tau = \min\{n_k\} / \max\{n_k\}$, where $n_k$ denotes the number of training samples in the $k$-th class. We fix $\max\{n_k\} = 16$ across all settings.
\section{Ablative Analysis}
\subsection{Ablation on the Number of Samples per Cluster.}
\begin{figure}[t]
\centering{
\includegraphics[width=1\linewidth]{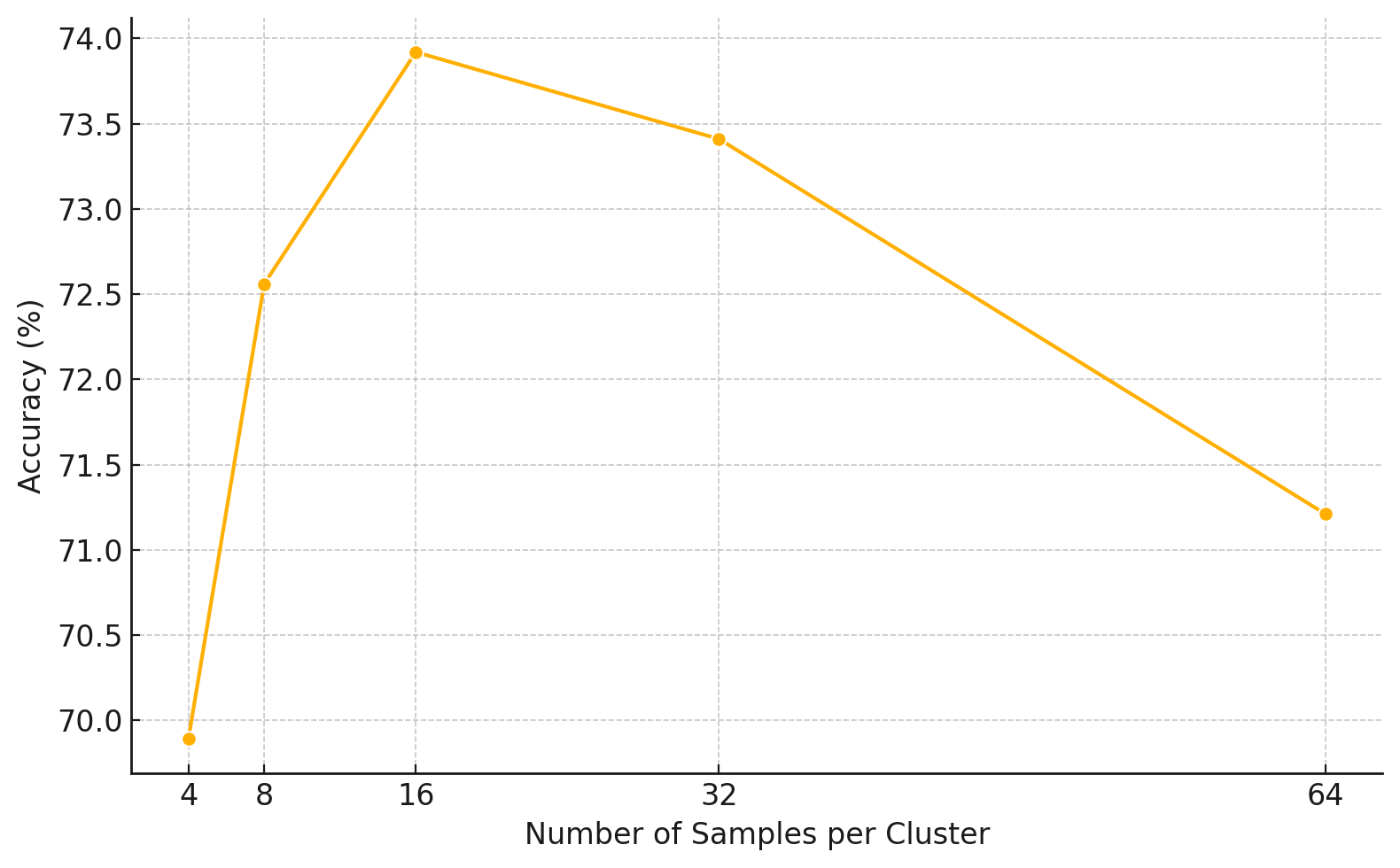}}
\caption{
Effect of the number of samples per cluster on base-to-new generalization performance.
Results are reported on the ImageNet dataset, where each point represents the harmonic mean (\textbf{H}) between base and novel class accuracy.
}
\label{fig:cluster_sensitivity}
\end{figure}

To investigate the sensitivity of our method to the number of samples per cluster, we conduct an ablation study by varying this value across a range of settings. This analysis explores how the granularity of semantic grouping—measured by how many class prototypes are grouped into each cluster—affects model performance and stability.
In our framework, frozen textual prototypes are grouped into disjoint semantic clusters using K-means before training. The number of samples per cluster determines how coarse or fine the semantic partitioning is. Larger clusters (i.e., more samples per cluster) result in coarser partitions, enabling broader semantic sharing but potentially reducing local discriminability. In contrast, smaller clusters offer finer-grained separation, which may improve within-cluster alignment but risks overfitting due to reduced intra-cluster diversity.
To better understand this trade-off, we evaluate CPT under varying cluster sizes, specifically using average per-cluster sample counts of $\{4, 8, 16, 32, 64\}$ prototypes. Results in Table~\ref{fig:cluster_sensitivity} show that performance improves as cluster size decreases, peaking at 73.92\% when each cluster contains 16 prototypes.
\subsection{Ablation on the choice of clustering algorithm.}
Our scaffold (Sec.~3.2) performs \emph{static} clustering on frozen textual features to preserve the pre-trained semantic structure and to provide cluster-local neighborhoods for NC constraints.
While the main paper uses K-means for static cluster mining, we examine whether alternative clustering algorithms affect the scaffold quality and the final performance, as shown in Table\ref{tab:cluster_ablation}.
We evaluate four common algorithms in the ImageNet base-to-new setting:
(1) Euclidean K-means, (2) Cosine K-means (spherical), (3) Spectral clustering (cosine affinity, normalized Laplacian), and (4) K-medoids (PAM, cosine distance).
For fairness, we fix the cluster count $M$ across algorithms, use the \emph{same} random-seed grid, and tune algorithm-specific hyperparameters \emph{only} on the frozen features via a cosine-silhouette criterion (no label or test data is used).
We report two metrics: (i) base-to-new harmonic mean (H) under $\tau\!\in\!\{0.25,0.06\}$ and (ii) seed-to-seed standard deviation (stability).
Across datasets we observe three consistent trends.
First, cosine-consistent partitioners (Cosine K-means) achieve the best H on both imbalance levels, with lower run-to-run variance than Euclidean K-means.
Second, Spectral clustering is competitive but more sensitive to initialization and the affinity scale, yielding slightly lower H and higher variance.
Third, K-medoids tends to underperform in both H and stability, indicating that medoid-only representatives provide a weaker scaffold for our cluster-local NC losses.
These results support our choice of a cosine-aligned partitioner for the static scaffold while confirming that the gains of CPT do not rely on a specific clustering heuristic.

\end{document}